\documentclass[runningheads]{llncs}
\usepackage[T1]{fontenc}
\usepackage{graphicx}
\usepackage[table]{xcolor}

\usepackage{float}

\usepackage{etoolbox}
\apptocmd{\sloppy}{\hbadness 10000\relax}{}{}

\usepackage{graphicx}

\usepackage{hyperref}
\usepackage{cleveref}

\usepackage[numbers]{natbib}

\usepackage[nolist]{acronym}
\begin{acronym}
    \acro{AURORA}{A Multicenter Analysis of Stereotactic Radiotherapy to the
        Resection Cavity of Brain Metastases}
    \acro{BLS}{BrainLesion Suite}
    \acro{BLP}{BrainLesion Suite preprocessing}
    \acro{BraTS}{Brain Tumor Segmentation}
    \acro{BTK}{BraTS Toolkit}
    \acro{CNN}{Convolutional Neural Network}
    \acro{DL}{Deep Learning}
    \acro{DSC}{Dice similarity coefficient}
    \acro{DQE}{Deep Quality Estimation}
    \acro{FL}{Federated Learning}
    \acro{GAN}{Generative Adversarial Network}
    \acro{GUI}{Graphical User Interface}
    \acro{IoU}{Intersection over Union}
    \acro{ML}{Machine Learning}
    \acro{MRI}{Magnetic Resonance Imaging}
    \acro{PGT}{Peak Ground Truth}
    \acro{RWMP}{Real World Model Performance}
    \acro{NN}{Neural Network}
    \acro{mpMRI}{multi-parametric MRI}
    \acro{T1n}{native T1-weighted}
    \acro{T1c}{contrast enhanced T1-weighted}
    \acro{T2w}{T2-weighted}
    \acro{T2f}{T2-weighted fluid-attenuated inversion recovery}
    \acro{ISLES}{Ischemic Stroke Lesion Segmentation}
    \acro{MS}{Multiple Sclerosis}
    \acro{AIMS-TBI}{automated identification of moderate-severe traumatic brain injury lesions}
    \acro{HIE}{Hypoxic Ischemic Encephalopathy}
    \acro{CaPTk}{Cancer Imaging Phenomics Toolkit}
    \acro{HD-GLIO}{HD-GLIO brain tumor segmentation tool}
    \acro{DeepMedic}{Efficient Multi-Scale 3D Convolutional Neural Network for Medical Image Segmentation}
    \acro{Panoptica}{Instance-wise Evaluation of Semantic Model
        Predictions}
    \acro{PQ}{Panoptic Quality}
    \acro{ASSD}{Average Symmetric Surface Distance}
    \acro{clDSC}{Centerline Dice}
    \acro{SQ}{Segmentation Quality}
    \acro{RQ}{Recognition Quality}
    \acro{QC}{quality control}
    \acro{ROI}{regions of interest}
    \acro{GBM}{Glioblastoma Multiforme}
    \acro{CSF}{cerebrospinal fluid}
    \acro{PBTs}{Pediatric brain tumors}
    \acro{T2H}{T2-hyperintense region}
    \acro{ET}{Enhancing tumor}
    \acro{CC}{Cystic component}
    \acro{WT}{Whole tumor }
\end{acronym}

\begin{document}
%
% samethanks for author list
\newcommand*\samethanks[1][\value{footnote}]{\footnotemark[#1]}

\newcommand{\eac}[1]{\emph{\ac{#1}}}
\newcommand{\eacp}[1]{\emph{\acp{#1}}}
\newcommand{\eacf}[1]{\emph{\acf{#1}}}
\newcommand{\eacfp}[1]{\emph{\acfp{#1}}}
\newcommand{\eacl}[1]{\emph{\acl{#1}}}

\title{\eacl{BLS}:
    \\ A Flexible and User-Friendly Framework for \\ Modular Brain Lesion Image Analysis}

\titlerunning{\eacl{BLS}}

\author{Florian Kofler\inst{1,2,3,4} \and
    Marcel Rosier \inst{1,2,8} \and
    Mehdi Astaraki \inst{12,13} \and
    Hendrik Möller \inst{4, 5} \and
    Ilhem Isra Mekki \inst{2} \and
    Josef A. Buchner \inst{6} \and
    Anton Schmick \inst{12} \and
    Arianna Pfiffer \inst{9} \and
    \\Eva Oswald \inst{2,7} \and
    Lucas Zimmer \inst{4} \and
    Ezequiel de la Rosa \inst{1} \and
    Sarthak Pati \inst{11,9} \and
    \\Julian Canisius \inst{3} \and
    Arianna Piffer \inst{15} \and
    Ujjwal Baid \inst{16} \and
    Mahyar Valizadeh \inst{2} \and
    \\Akis Linardos \inst{9} \and
    Jan C. Peeken \inst{6, 17, 18} \and
    Surprosanna Shit \inst{1} \and
    Felix Steinbauer \inst{8} \and
    Daniel Rueckert \inst{9} \and
    Rolf Heckemann \inst{9} \and
    Spyridon Bakas \inst{9} \and
    Jan Kirschke \inst{9} \and
    \\Constantin von See \inst{10} \and
    Ivan Ezhov \inst{9} \and
    \\Marie Piraud \inst{2} \thanks{equal contribution} \and
    Benedikt Wiestler \inst{3} \samethanks \and
    Bjoern Menze \inst{1} \samethanks
}

% Isra Mekki (HAI)
% Mahyar Valizadeh (HAI)
% Stefan Ehrlich
% Ivan Ezhov (TUM)
% Suprosanna Shit (UZH)

% \samethanks
%

\authorrunning{F. Kofler et al.}

\institute{
    % 1
    Department of Quantitative Biomedicine, University of Zurich, Zurich, Switzerland\and
    % 2
    Helmholtz AI, Helmholtz Zentrum München, Germany \and
    % 3
    Department of Diagnostic and Interventional Neuroradiology, School of Medicine, Klinikum rechts der Isar, Technical University of Munich, Germany \and
    % 4
    TranslaTUM - Central Institute for Translational Cancer Research, Technical University of Munich, Germany \and
    % 5
    Chair for AI in Medicine, Klinikum Rechts der Isar, Munich, Germany \and
    % 6
    Department of Radiation Oncology, Klinikum rechts der Isar, Technical University of Munich, Germany \and
    % 7
    Institute of Clinical Neuroimmunology, Faculty of Medicine, Ludwig Maximilian University, Munich, Germany \and
    % 8
    School of Computation, Information and Technology, Technical University of Munich, Germany \and
    % 9
    Division of Computational Pathology, Indiana School of Medicine, Indianapolis, IN, USA\and
    %10
    Research Center for Digital Technologies in Dentistry and CAD/CAM, Department of Dentistry, Faculty of Medicine and Dentistry, Danube Private University, Krems, Austria \and
    % 11
    Medical Research Group, MLCommons, San Francisco, CA, USA
    \and
    % 11
    Department of Neurology, Clinical Neuroscience Center and Brain Tumor Center, University and University Hospital Zurich, 8091 Zurich, Switzerland
    \and
    %12
    Department of Medical Radiation Physics, Stockholm University, Solna, Sweden
    \and
    %13
    Department of Oncology-Pathology, Karolinska Institutet, Solna, Sweden, Sweden
    \and
    %14
    Division of Oncology and Children's Research Center, University Children's Hospital Zurich, Switzerland
    \and
    %15
    Wallace H. Coulter Department of Biomedical Engineering at Georgia Tech and Emory University, Atlanta, USA;
    \and
    %16
    Institute of Radiation Medicine (IRM), Helmholtz Zentrum München (HMGU), Germany
    \and
    %17
    German Consortium for Translational Cancer Research (DKTK), Partner Site Munich, Munich, Germany
}

\maketitle              % typeset the header of the contribution
\begin{abstract}
    \eacl{BLS} is a versatile toolkit for building modular brain lesion image analysis pipelines in Python.
    Following Pythonic principles, \eacl{BLS} is designed to provide a 'brainless' development experience, minimizing cognitive effort and streamlining the creation of complex workflows for clinical and scientific practice.
    At its core is an adaptable preprocessing module that performs co-registration, atlas registration, and optional skull-stripping and defacing on arbitrary multi-modal input images.
    \eacl{BLS} leverages algorithms from the BraTS challenge to synthesize missing modalities, inpaint lesions, and generate pathology-specific tumor segmentations.
    \eacl{BLS} also enables quantifying segmentation model performance, with tools such as panoptica to compute lesion-wise metrics.
    Although \eacl{BLS} was originally developed for image analysis pipelines of brain lesions such as glioma, metastasis, and multiple sclerosis, it can be adapted for other biomedical image analysis applications.
    The individual \eacl{BLS} packages and tutorials are accessible on \href{https://github.com/BrainLesion/}{GitHub}.

\end{abstract}

\section{Introduction}
\label{sec:introduction}
% \eac{DL} has revolutionized biomedical image analysis.
% Hundreds of papers demonstrate how \eac{DL} methods can be used across various tasks, including image registration, classification, object detection, and semantic and instance segmentation.
% However, most of these innovations never find their way into clinical and scientific practice.

\eac{MRI} is the gold standard modality for non-invasive investigation of structural and functional characteristics of the brain \citep{MRIforBrain}.
\eac{MRI} has become essential in the diagnosis and treatment planning for a wide range of neurological disorders, including neurodegenerative diseases (such as dementia )\citep{MRIappDementia}, neuroimmunological diseases (such as multiple sclerosis, among others) \citep{MRIappMS}, and tumors \citep{MRIappTumor}.
However, the inherent complexity of high dimensional \eac{mpMRI} volumes renders manual interpretation and subjective measurement not only laborious but also susceptible to inter- and intra-observer variability.
Therefore, significant research efforts have been conducted towards the development of automated and/or semi-automated computational methods for brain \eac{MRI} analysis.

\eac{DL} methodologies have significantly transformed biomedical image analysis.
Extensive research has demonstrated the effectiveness of \eac{DL} across a range of tasks, including image registration \citep{TransMorphRegistration}, anomaly detection \citep{BauerAnomaly}, classification \citep{ClassificationTransfomers}, object detection \citep{nnDetectionDKFZ}, semantic \citep{Isensee2021nnUnet,
    IsenseennUnetRevisited} and instance segmentation \citep{Moller2024Spine}.
The introduced robust algorithms have facilitated the development of numerous imaging biomarkers for critical diagnostic \citep{ASTARAKILungCancer,
    DLRapidHistology} and prognosis \citep{Anahita24GenomeMRI} \citep{Bakas2017GenomeAtlas} applications.
However, translating these methodological advancements into clinical and scientific practice remains challenging due to several factors.
These include, but are not limited to, inter-modality variations (e.g., \textit{CT}, \textit{MRI}, \textit{PET}, multimodal), inconsistent target definitions (e.g., lytic vs. sclerotic metastasis segmentation), variations in data preparation and preprocessing, and a lack of standardized performance evaluation metrics (e.g., lesion-wise vs. subject-wise analysis).

To address these inconsistencies, international benchmarking competitions have become essential for comparative performance assessment in biomedical image analysis.
These challenges target clinically relevant problems by establishing standardized frameworks for objective algorithm evaluation on common datasets.
This allows for the identification of strengths and weaknesses in current methodologies.
In the context of brain \eac{MRI}, several prominent challenges have emerged.
The \eac{BraTS} \citep{BraTS2015Menze}, a leading force in \eac{MRI}-based tumor analysis since 2012, initially focused on glioma segmentation in \eac{mpMRI} including \eac{T1n}, \eac{T1c}, \eac{T2w}, and \eac{T2f} images \citep{bakas2019identifyingbestmachinelearning}.
\eac{BraTS} has since expanded its scope to encompass diverse tumor types and applications beyond segmentation, such as histological image analysis \citep{bakas2024bratspathchallengeassessingheterogeneous}.
Other significant challenges include the \eac{ISLES} challenge for stroke lesion segmentation in structural and diffusion-weighted \eac{MRI} \citep{ISLESHernandezPetzsche2022}, the \eac{MS} lesion segmentation challenge for \eac{MS} lesion detection and segmentation in \eac{mpMRI} \citep{MSSEG2OlivierCommowick}, the \eac{AIMS-TBI} challenge for \eac{T1n} \eac{MRI} lesion segmentation, and the more recent \eac{HIE} challenge focusing on \eac{HIE}-related lesion segmentation and outcome prediction in infant brain \eac{MRI} \citep{HIESeg}.
A recent meta-analysis of 80 challenges indicated that 43\% of winning algorithms surpassed state-of-the-art methods, with 11\% achieving complete problem-domain solutions \citep{WhyChallengeWinnerCVPR}.

Despite the promising results from these benchmarking challenges, it is crucial to recognize that challenge datasets typically undergo rigorous anonymization, standardized preprocessing, and extensive quality control for both images and reference labels.
Consequently, the usability and generalizability of developed models on private datasets depend on replicating these standardized processing pipelines.
This necessity underscores the importance of developing a modular pipeline that is easily executable, integrable with other frameworks, and deployable across diverse applications.

\textbf{Related work:} Several tools have been developed to address the challenges of brain lesion segmentation and analysis.

The \eac{CaPTk} \citep{pati2020cancer} is a versatile platform designed for neuro-oncological research, offering tools for segmentation, feature extraction, and predictive modeling, with a focus on brain tumors, including gliomas and metastases, although it can be computationally demanding and complex to install.
\eac{HD-GLIO} \citep{kickingereder2019automated} explicitly focuses on glioma segmentation, providing a streamlined pipeline that integrates both preprocessing and segmentation, yet it has limited flexibility with input modalities.
\eac{BraTS} Toolkit \citep{kofler2020brats} is known for its user-friendly and modular design, refined through extensive benchmarking in the \eac{BraTS} challenges, excelling in preprocessing, segmentation, and the fusion of segmentations derived from \eac{BraTS} challenge algorithms, particularly for gliomas, but also applicable to other brain lesions.
\eac{DeepMedic} \citep{kamnitsas2016deepmedic} employs a 3D convolutional architecture to achieve high accuracy in brain lesion segmentation, addressing various conditions such as gliomas, strokes, and traumatic brain injuries, although it necessitates external preprocessing and substantial computational resources.

\textbf{Contribution:} Building on the strengths and limitations of these existing tools, the \eac{BLS} Toolkit aims to provide a 'brainless' development experience for building modular brain lesion image analysis pipelines.
With an adaptable preprocessing module that handles co-registration, atlas registration, defacing, N4 bias field corrections, \eac{BLS} expands its capabilities beyond gliomas to include various brain lesions, making it a flexible solution for diverse clinical and scientific applications.
By leveraging a range of algorithms, including those from the \eac{BraTS} challenge, and enabling the quantification of model performance, \eac{BLS} seeks to streamline the creation of complex workflows in brain image analysis.

\section{BrainLesion Suite}
\label{sec:modules}
The \eac{BLS} consists of multiple modules, most of which are hosted under its \href{https://github.com/BrainLesion}{GitHub organization}.
The modular structure enables the creation of powerful bio-medical image analysis pipelines by combining \eac{BLS} and non-\eac{BLS} modules.
% TODO a figure would be nice here
% TODO it would be nice to have a good example of an already published paper here
\eac{BLS} follows the philosophy of open and reproducible science; therefore, whenever possible, \eac{BLS} modules are released under the \href{https://www.apache.org/licenses/LICENSE-2.0}{Apache License, Version 2.0}.
% Nevertheless, besides Linux, \eac{BLS} packages, make an effort to support proprietary operating systems, such as Microsoft Windows and Mac OSX, whenever possible to embrace the sometimes harsh realities in clinical and research institutions.
More importantly, \eac{BLS} package development strategy prioritizes cross-platform compatibility beyond Linux, extending to proprietary operating systems such as Microsoft Windows and macOS.
This approach aims to address the diverse and often constrained technological environments prevalent in clinical and research institutions.
The functionality of the respective modules is described below.

\subsection{Multi-modal image to atlas preprocessing}
\label{sec:preprocessing}

% The preprocessing module can co-register multi-modal images, register them to an atlas of choice, such as the integrated SRI-24 \citep{rohlfing2010sri24}, and provide brain extraction, defacing, N4-bias field correction, and normalization functionality.
% To achieve these steps it tries to make use of well-established methods wherever possible.
% Nevertheless, the preprocessor module can easily be extended with new defacing, skull stripping, normalization, and registration methods due to its \emph{backend agnostic} design.
% At the time of writing, the following methods are implemented:

% For registrations we support \href{https://github.com/ANTsX/ANTsPy}{ANTsPy} \citep{avants2009advanced} and \href{https://github.com/KCL-BMEIS/niftyreg}{niftyreg} \citep{ourselin2001reconstructing, modat2014global}.
% Even though the preprocessing module can be run with custom sophisticated registration scripts, we recommend conducting only rigid registrations to avoid distorting volumetry calculations.
% Furthermore, HD-bet \citep{isensee2019automated} provides the brain extraction functionality and
% defacing is implemented via Quickshear \citep{schimke2011quickshear}.
% N4 bias field correction is implemented via \href{https://simpleitk.readthedocs.io/en/master/link_N4BiasFieldCorrection_docs.html}{sitk} \citep{tustison2010n4itk}.

The preprocessing module facilitates the co-registration of multi-modal images, registration to user-defined atlases (such as integrated SRI-24 \citep{rohlfing2010sri24}, and MNI-152 \citep{MniAtlas2009Paper} or any NIfTI-formatted brain \eac{MRI} atlas).
Additional functionalities include brain extraction, defacing, N4 bias field correction, and intensity normalization.
The described steps were implemented by leveraging established methods, whenever possible. Nevertheless, the module's architecture is designed for extensibility, allowing for the integration of alternative algorithms for defacing, skull stripping, intensity normalization, and registration.

Currently, registration is supported through \href{https://github.com/ANTsX/ANTsPy}{ANTsPy} \citep{avants2009advanced} and \href{https://github.com/KCL-BMEIS/niftyreg}{niftyreg} \citep{ourselin2001reconstructing, modat2014global}, \href{https://elastix.dev/}{elastix} \citep{elastix_0,elastix_1}, and \href{https://github.com/pyushkevich/greedy}{greedy} \citep{greedy}.
While sophisticated custom registration scripts can be incorporated, rigid registration is recommended to avoid potential volumetric distortions.
Brain extraction is performed using HD-bet \citep{isensee2019automated}, and defacing is implemented via Quickshear \citep{schimke2011quickshear}.
Different intensity normalization strategies are available and \emph{N4 bias field correction} is implemented using \href{https://simpleitk.readthedocs.io/en/master/link_N4BiasFieldCorrection_docs.html}{sitk} \citep{tustison2010n4itk}.

\begin{figure}[hbtp]
    \centering
    ®
    \includegraphics[width=0.7\linewidth]{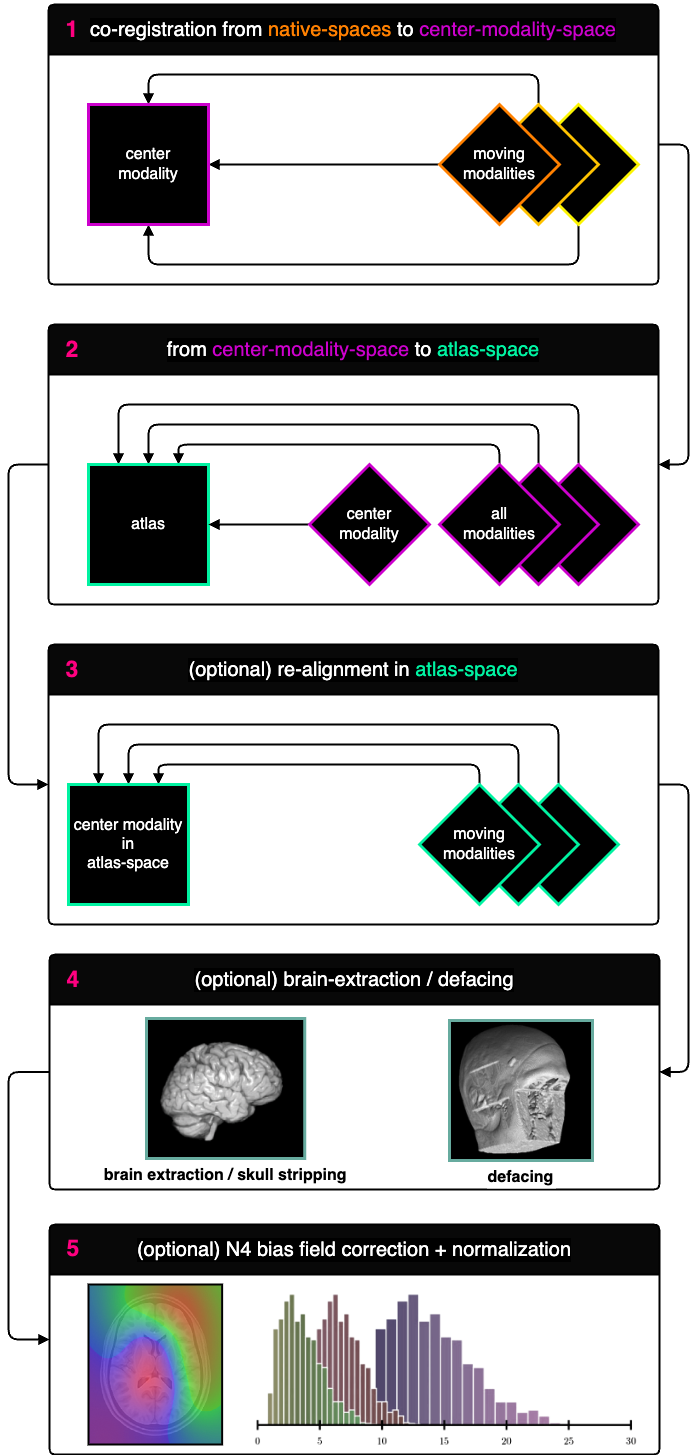}
    \caption{
        \emph{Multi-modal brain MRI preprocessing.}
        Squares indicate registration targets, diamonds moving images; outline colors encode varying image spaces.
        First, the moving modalities are registered from their respective native spaces to the center modality in its native space.
        Second, the center modality is registered to the target atlas, and the remaining moving modalities are transformed into atlas space using the resulting transformation matrix.
        The following three steps are optional:
        Third, the moving modalities are again co-registered to the center modality within the atlas space to minimize registration inaccuracies.
        Fourth, the brain MRIs are skullstripped and/or defaced.
        Last, N4 bias field correction and image intensity normalization can be applied.
        The preprocessing module is back-end agnostic, meaning it supports conducting the registrations, brain extraction, and defacing with different methods.
    }
    \label{fig:preprocessor}
\end{figure}

\autoref{fig:preprocessor} provides a schematic representation of the \eac{BLP} module's workflow, illustrating its role in enabling multimodal brain \eac{MRI} lesion analysis pipelines.
User guide and source code are available on the \href{https://github.com/BrainLesion/preprocessing}{\eac{BLP} GitHub repository}

% \Cref{fig:preprocessor} illustrates the \eac{BLP}'s mechanics of the multi-modal image preprocessing module meant to enable multi-modal brain \eac{MRI} lesion analysis pipelines.
% User guide and source code are available on the \href{https://github.com/BrainLesion/preprocessing}{\eac{BLP} GitHub repository}.

% \textbf{\textcolor{red}{from SP: does it make sense to talk about eReg here?}}

\subsection{Multi-modal MR sequence identification with modsort}
\label{sec:modsort}

\begin{figure}[H]
    \centering
    \includegraphics[width=1.0\linewidth]{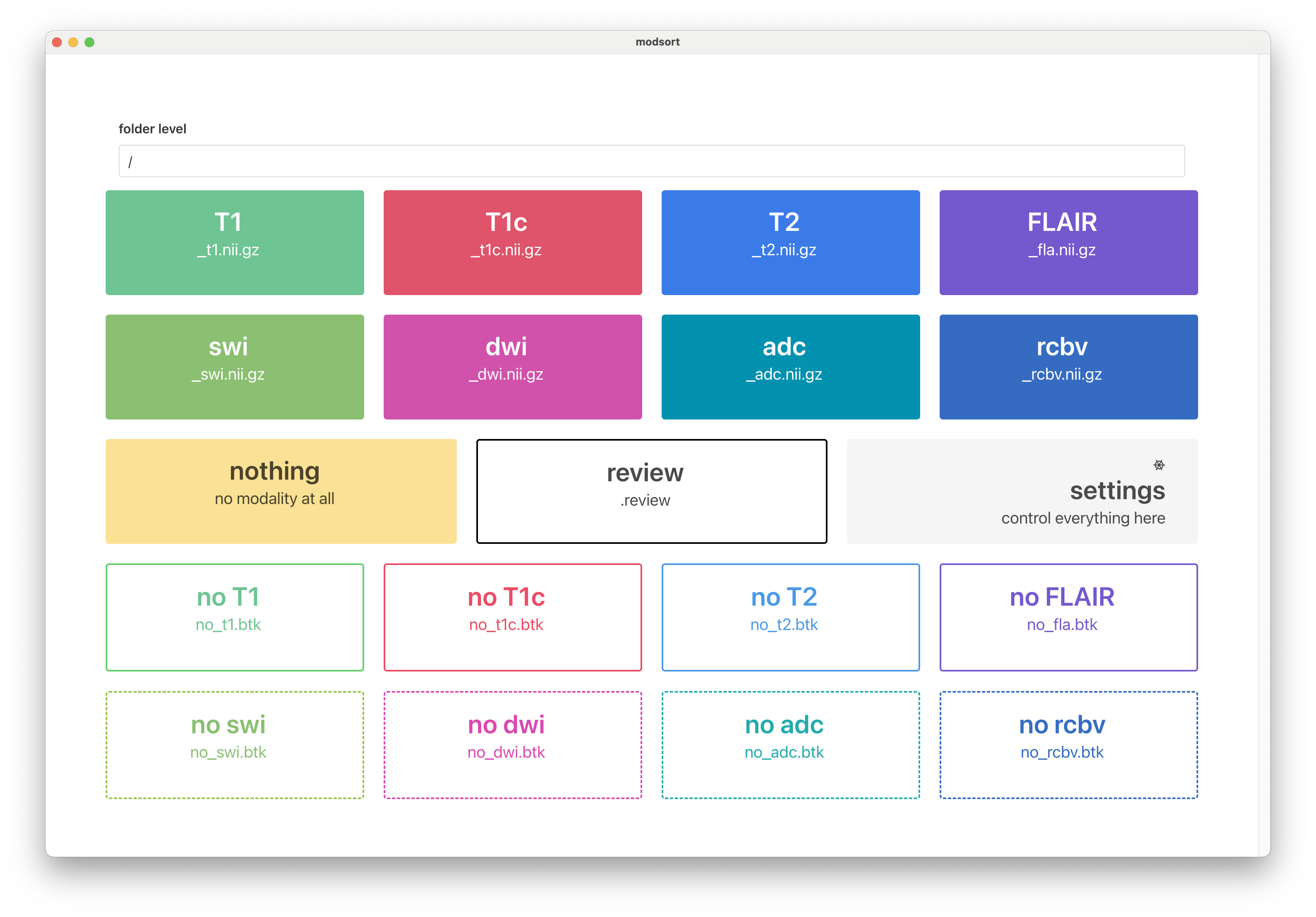}
    \caption{
        The \emph{modsort} \eac{GUI}.
        Users can drag and drop MR files into labeled boxes to manually sort and tag sequences.
        Files are automatically renamed and structured, reducing errors and streamlining preprocessing.
    }
    \label{fig:modsort_gui}
\end{figure}

\href{https://github.com/BrainLesion/modsort}{Modsort}
is an interactive tool for sorting, identifying, and tagging \eac{MRI} sequences.
It enables users to manually classify medical imaging files using an intuitive drag-and-drop interface (see \autoref{fig:modsort_gui}).

Files are automatically renamed, structured, and copied to a relative location, ensuring consistency and eliminating errors.
A bulk mode allows efficient renaming of multiple files at once.
\textit{Modsort} is particularly useful for organizing \eac{MRI} datasets before preprocessing with the \eac{BLS}, improving workflow efficiency for clinical and research applications.
\textit{Modsort} is available on \href{https://github.com/BrainLesion/modsort}{GitHub}.

\subsection{BraTS orchestrator: - Algorithms from the \eac{BraTS} challenge}
\label{sec:brats}

\begin{figure}
    \centering
    \includegraphics[width=\linewidth]{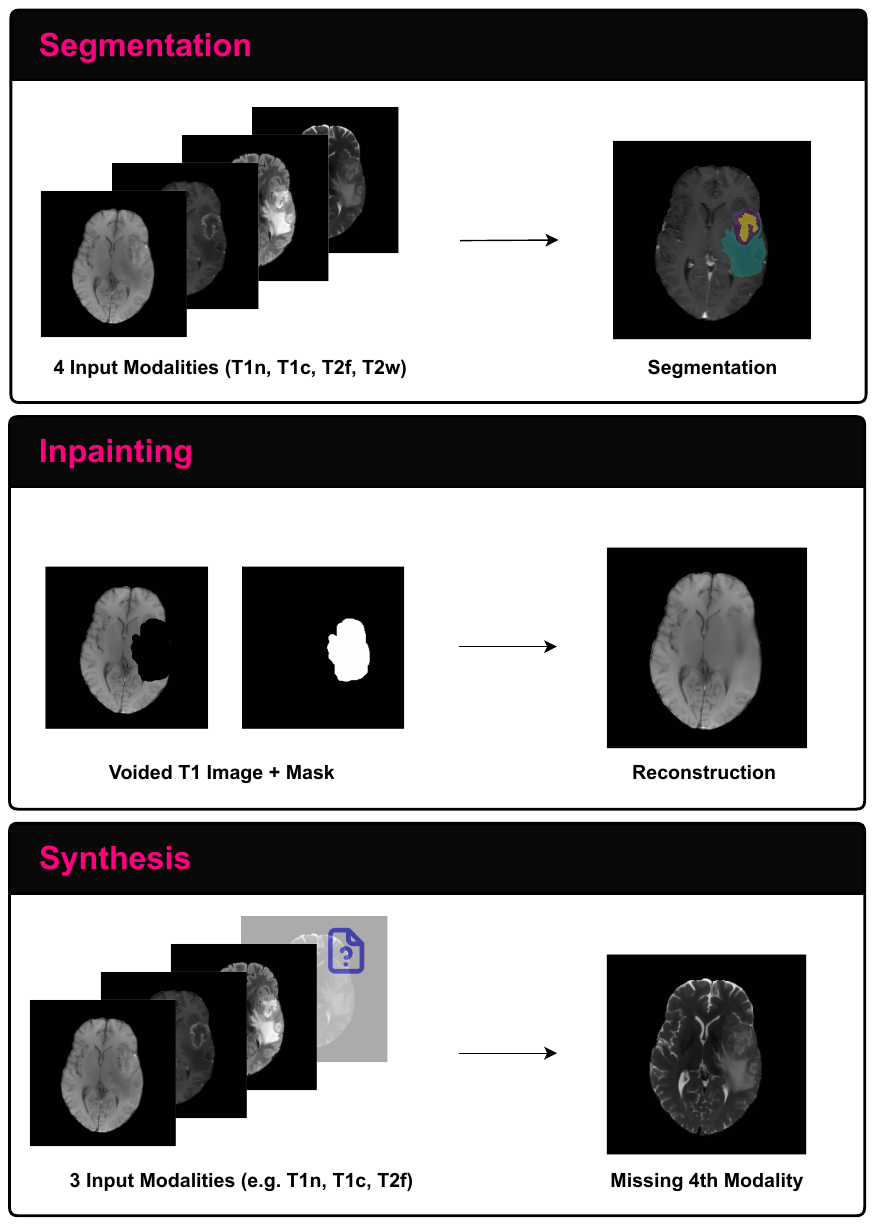}
    \caption{Overview visualizing the different algorithm types available in the BraTS package with their respective inputs and outputs.}
    \label{fig:brats_tasks}
\end{figure}

The annual \eac{BraTS} challenge has, in recent years, fostered the development of innovative solutions to clinically relevant problems across different brain tumor entities, addressing relevant clinical needs.
Prominent task categories within \eac{BraTS} include various tumor segmentation tasks (e.g., glioma, meningioma, metastasis, pediatric) \citep{
    adewole2023braintumorsegmentationbratsafrica,
    baid2021rsnaasnrmiccaibrats2021benchmark,
    deverdier20242024braintumorsegmentationadultgliomaposttreatment,
    kazerooni2024braintumorsegmentationpediatrics24,
    kazerooni2024braintumorsegmentationbratsped23,
    labella2023asnrmiccaibraintumorsegmentationMeningioma,
    labella2024braintumorsegmentationbratsMeningioma,
    moawad2024braintumorsegmentationbratsmets}, lesion inpainting task for the reconstruction of voided brain regions \citep{kofler2024braintumorsegmentationbratsinpainting}, and synthesizing missing \eac{MRI} sequence task, which involves generating a missing \eac{MRI} sequence from a set of available sequences of the same subject (e.g., synthesizing \eac{T2f} from \eac{T1c}, \eac{T1n}, and \eac{T2w} images) \citep{li2024braintumorsegmentationbratsbrasyn}.
\autoref{fig:brats_tasks} provides a visual representation of these tasks and their respective input and output data.

% The \href{https://github.com/BrainLesion/BraTS/}{BraTS package} is an effort to make the best submissions from the discussed challenges available to end users via an easy-to-use Python API powered by docker.
% It continues and simplifies the segmentation module of \citep{kofler2020brats} that similarly provided algorithms from older challenges.

The \href{https://github.com/BrainLesion/BraTS/}{\eac{BraTS} orchestrator} \citep{kofler2025bratsorchestratordemocratizing} is an effort to disseminate top-performing models, developed for each task within the \eac{BraTS} challenge, to end users through a user-friendly Python API facilitated by Docker containerization.
This package constitutes a streamlined successor to the prior segmentation module, the \eac{BraTS} toolkit \citep{kofler2020brats}, which served a comparable function in providing access to top-performing algorithms from preceding \eac{BraTS} challenges.

\subsection{AURORA: Adaptive modality-aware metastasis segmentation}
\label{sec:aurora}

Brain metastases are approximately ten times more common than primary brain tumors like gliomas \citep{mitchell2022brain}.
Treatment strategies depend on the performance status of the patients and the characteristics of the lesions: Stereotactic radiation therapy is typically used in patients with a limited number of metastases \citep{rogers2022stereotactic}, while larger singular lesions may require surgical resection \citep{vogelbaum2022treatment}.
Treatment recommendations are formulated in interdisciplinary tumor boards.
After resection, the cavity may be irradiated to improve local control rates \citep{minniti2017outcomes}.

Accurate delineation of the metastasis is essential for treatment planning and outcome prediction \citep{buchner2024radiomics}.
Although \eac{T2f}-hyperintense edema surrounding brain metastases is not routinely included in the delineation of the radiotherapy target, in gliomas some guidelines recommend its inclusion \citep{kruser2019nrg}.
Additionally, the characteristics of surrounding \eac{T2f}-hyperintense edema can aid in distinguishing metastases from gliomas \citep{priya2021machine}.

\eac{DL} models for automated segmentation of brain metastases and their surrounding \eac{T2f}-hyperintense edema \citep{buchner2023development, buchner2023identifying} were developed as part of the \eac{AURORA} multicenter study.
The original model \citep{buchner2023development} leverages four \eac{MRI} sequences, following the \eac{BraTS} challenge framework.
As part of an ablation study, additional models have been developed to test the effectiveness of lowering the requirements---i.e. leveraging fewer input sequences \citep{buchner2023identifying}.
% The \href{https://github.com/BrainLesion/AURORA/}{models} are provided as part of the \eacf{BLS} to allow easy segmentation of brain metastases and their surrounding edema prior to resection. 

As input, \eac{AURORA} requires skull-stripped \eac{MRI} sequences in 1 millimeter isotropic resolution.
The \eac{BLP} module ensures standardized input preparation. The models have been validated on large-scale multicenter datasets, demonstrating state-of-the-art segmentation performance \citep{buchner2023development}.

The \eac{AURORA} framework presents a significant step toward automated, accurate, and efficient segmentation of brain metastases in clinical and research settings.
The software, available as an \href{https://github.com/BrainLesion/AURORA}{open-source implementation}, supports integration into existing neuro-oncology workflows.
% The appropriate models can be chosen depending on the available sequences and labels of interest.

% The \eacf{BLS} preprocessing module is recommended for this purpose.

\subsection{GlioMODA: Adaptive modality-aware glioma segmentation}

Gliomas are the most common group of primary brain tumors in adults, consisting of heterogeneous tumors all arising from glial (progenitor) cells.
\eac{GBM} represents the most aggressive subtype, characterized by diffuse infiltration into surrounding brain tissue and rapid progression \citep{Wen2008,Louis2021}.
Since the 2016 edition of the WHO classification, gliomas have been classified based not only on histopathologic appearance but also on well-established molecular parameters \citep{Louis2016}.
Treatment strategies for gliomas differ by tumor histology and typically involve a multidisciplinary approach combining the major treatment modalities of surgery, radiotherapy, and systemic pharmacotherapy \citep{Weller2021}.
In addition to these established modalities, recent advances include molecular targeted therapies, immunotherapies, and other innovative approaches currently under investigation \citep{Tang2025}.

Accurate segmentation of glioma subregions is critical for treatment planning, prognosis assessment, and monitoring therapeutic response \citep{Sun2019,Niyazi2023}.
Manual segmentation on \eac{mpMRI} is particularly challenging due to the complex and heterogeneous structure of gliomas, including an enhancing core, central necrosis, and extensive surrounding edema \citep{Sun2019}.
\eac{MRI} modalities, including \eac{T1c} for enhancing tumor, \eac{T2w}, and especially \eac{T2f} sequences for edema, are widely used for their complementary imaging characteristics \citep{Ellingson2015}.
Missing or corrupted \eac{MRI} sequences present a common challenge in clinical routine, often due to factors like varying institutional protocols or artifacts such as patient motion.
The prevalence of this issue can be significant; for example, one study reported that over 20\% of sequences were missing from a large, multi-centric glioma dataset \citep{Pemberton2023}.

To address these limitations and promote the clinical application of glioma segmentation,\textit{GlioMODA} offers a collection of open-source, fully automated deep learning models designed for segmenting adult gliomas on \eac{mpMRI}, capable of handling varying input sequences.
Leveraging the robust nnU-Net architecture,\textit{GlioMODA} dynamically adapts to any available combination of the four standard MRI modalities (\eac{T1c}, \eac{T1n}, \eac{T2w}, \eac{T2f}).
The workflow includes input validation, automated modality detection, appropriate model selection, and seamless management of pretrained weights.
The model produces four output channels corresponding to: (1) background, (2) necrotic and non-enhancing tumor core, (3) edema, and (4) enhancing tumor, in line with established \eac{BraTS} segmentation conventions \citep{Kofler2020}.
\textit{GlioMODA} has been validated on large, multicenter datasets, demonstrating state-of-the-art accuracy.
The Python package is available on \href{https://github.com/BrainLesion/GlioMODA}{GitHub} for straightforward integration into clinical and research pipelines, facilitating reproducible and efficient brain tumor segmentation.

% link to package

% TODO link:
% https://github.com/BrainLesion/GlioMODA

% * similar to AURORA but nnUnet instead of MONAI backend
% * explain output channels

\subsection{PeTu: Adaptive modality-aware pediatric tumor segmentation}
\label{sec:pediatric}
% TODO

% link to package

% TODO link:
% https://github.com/BrainLesion/PeTu

% * similar to AURORA but nnUnet instead of MONAI backend
% * explain output channels

\eac{PBTs} are the most common solid tumors in children and represent the leading cause of cancer-related death in the pediatric population.
Treatment strategies vary considerably depending on tumor histology, location, and extent of spread, and often require a combination of surgical resection, radiotherapy, and chemotherapy.
Accurate, consistent, and timely delineation of tumor subregions is critical for treatment planning, prognosis estimation, and longitudinal monitoring.

\eac{DL} models have demonstrated significant promise in recent years for automated tumor segmentation across diverse imaging modalities and tumor entities \citep{Decathlon2022}.
Among the many robust segmentation pipelines developed, nnU-Net \citep{Isensee2021nnUnet} stands out as a self-configuring robust \eac{DL} framework.
It automatically adapts to the characteristics of a given dataset, eliminating the need for manual hyperparameter tuning.

This study introduces a fully automated segmentation pipeline for \eac{PBTs}, leveraging a 3D nnU-Net framework trained on co-registered \eac{mpMRI} data (\eac{T1c}, \eac{T1n}, \eac{T2w}, \eac{T2f}).
The nnU-Net pipeline was selected due to its demonstrated high performance across diverse medical image segmentation tasks \citep{Isensee2021nnUnet}.
The model was trained for 1000 epochs across five folds, utilizing default configurations, with 250 training iterations and 50 validation iterations per epoch. An initial learning rate of 1e-2 and a batch size of 2 were employed, with deep supervision enabled.
The model accepts four-channels input tensors composed of co-registered \eac{mpMRI} scans and subsequently segments tumoral regions into their constituent subregions, including:

\begin{enumerate}
    \item \textbf{\eac{T2H}} – typically encompassing both the solid tumor mass and associated edema.
    \item \textbf{\eac{ET}} – regions with contrast uptake, indicative of active or aggressive tumor areas.
    \item \textbf{\eac{CC}} – fluid-filled regions often seen in certain pediatric tumor types.
    \item \textbf{\eac{WT}} – an aggregate mask including all visible tumor components.

\end{enumerate}

% Each output is encoded as a binary mask and can be combined or analyzed independently, depending on the clinical or research objective. 
The model demonstrated strong segmentation performance for the \eac{WT} and \eac{T2H} regions, achieving moderate performance for the \eac{ET}.
However, performance for the \eac{CC} remained limited.

This nnU-Net-based framework offers a robust, reproducible, and scalable tool for \eac{PBTs} segmentation with clinical relevance.
It represents a significant advancement toward standardized image analysis in pediatric neuro-oncology.

This model is available as an \href{https://github.com/BrainLesion/PeTu}{open-source Python package}, and supports integration into existing neuro-radiology and neuro-oncology workflows.

\subsection{DQE: Deep Quality Estimation through Human Surrogate Models}
\label{sec:dqe}

The evaluation of automated glioma segmentation models remains a challenging task, particularly when considering inter-rater variability among human experts.
To address this challenge, \citet{kofler2022deep} proposes a novel approach utilizing \eac{DL}-based human surrogate models to estimate segmentation quality.
This methodology leverages a neural network trained to predict the quality of glioma segmentations as perceived by human raters, thereby providing a surrogate metric that aligns with expert judgment.

Unlike conventional evaluation techniques that rely solely on volumetric overlap metrics such as the \eac{DSC} or \eac{IoU}, \eac{DQE} integrates a learned quality predictor to approximate human assessment more accurately.
Concretely, the \eac{DQE} model infers segmentation quality based on patterns it has learned from these human evaluations, effectively serving as a data-driven surrogate for human assessment.
This approach has potential applications in:
\begin{itemize}
    \item \textbf{Quality Monitoring During Inference}: Despite fully-automatic glioma segmentation models often outperforming human annotators on average, they can occasionally fail dramatically \citep{maier2018rankings, kofler2023approaching}. For clinical translation, \eac{DQE} can help detect and mitigate such failure cases in real-time.
    \item \textbf{Dataset Curation}: \eac{DQE} enables automatic identification of unreliable ground truth labels, which is particularly valuable in \eac{FL} settings where researchers lack direct access to annotations \citep{sheller2020federated, rieke2020future, pati2022federated}. By applying a threshold to the estimated quality scores, only trustworthy cases can be used for model training.
    \item \textbf{Optimization of Training Objectives}: The ability of \eac{DQE} to approximate non-differentiable human quality assessments with a differentiable \eac{CNN} suggests its potential for integration into model loss functions. This opens avenues for training segmentation models directly on quality-aware objectives.
\end{itemize}

The \eac{DQE} package implements this methodology within a minimal Python package, enabling seamless integration into existing segmentation workflows.
The framework is compatible with common \eac{DL} pipelines and provides pretrained models for direct application to glioma segmentation tasks.

Experimental results demonstrate that \eac{DQE}-based quality assessment correlates strongly with human expert ratings, outperforming traditional overlap-based metrics in capturing perceptual segmentation accuracy.
The introduction of human surrogate models for segmentation evaluation is a promising step toward improving the interpretability and reliability of automated glioma segmentation systems.

The \eac{DQE} framework is available as an \href{https://github.com/BrainLesion/deep_quality_estimation}{open-source package}, facilitating its adoption in both research and clinical applications.

% \citet{kofler2022deep} introduce the idea of estimating the quality of BraTS glioma segmentation models via human surrogate deep learning models. 
% This concept can arguably be useful for evaluation purposes or as part of a loss function during model training.
% The \href{https://github.com/BrainLesion/deep_quality_estimation}{DQE package} makes the proposed model accessible through a minimal Python package. 

\subsection{Panoptica: Instance-wise Evaluation of Semantic Model Predictions}
\label{sec:panoptica}

\eac{Panoptica} is a versatile, performance-optimized Python package designed for computing instance-wise segmentation quality metrics from 2D and 3D segmentation maps~\citep{kofler2023panoptica}0. Unlike traditional approaches that rely solely on Intersection over Union (IoU)-based \eac{PQ}, \eac{Panoptica} extends evaluation capabilities with additional metrics such as the \eac{ASSD} and \eac{clDSC}, providing a more nuanced understanding of model performance.

The framework operates through a modular, three-step process:
\begin{enumerate}
    \item \textbf{Instance Approximation}: Extracts individual instances from semantic segmentation outputs, addressing the lack of explicit instance labels in many biomedical datasets.
    \item \textbf{Instance Matching}: Matches predicted instances with reference annotations, handling common issues like label mismatches.
    \item \textbf{Instance Evaluation}: Computes a comprehensive set of metrics, including traditional overlap scores and advanced panoptic metrics such as \eac{RQ}, \eac{SQ}, and \eac{PQ}.
\end{enumerate}

We demonstrate the efficacy of Panoptica on diverse biomedical datasets, highlighting its ability to provide detailed instance-wise evaluations that are critical for accurately representing clinical tasks.
The nuanced insights offered by Panoptica's extensive metric suite enable a deeper understanding of model strengths and weaknesses, supporting more informed model development and evaluation.

Panoptica is open-source and available on \href{https://github.com/BrainLesion/panoptica/}{GitHub}.
It is implemented in Python, compatible with Python 3.10+, and accompanied by comprehensive documentation and tutorials.

% TODO Figure

%TODO Hendrik
% panoptica, a versatile and performance-
% optimized package designed for computing instance-wise segmentation
% quality metrics from 2D and 3D segmentation maps. panoptica over-
% comes the limitations of existing metrics and provides a modular frame-
% work that complements the original IoU-based Panoptic Quality with
% other metrics, such as the distance metric ASSD. panoptica employs a
% three-step metrics computation process to cover diverse use cases. We
% demonstrate the efficacy of panoptica on various real-world biomedical
% datasets, where an instance-wise evaluation is instrumental for an ac-
% curate representation of the underlying clinical task, and the nuanced
% evaluation offered by panoptica provides important insights into model
% performance. The package is open-source, implemented in Python, and
% accompanied by comprehensive documentation and tutorials. Overall, we
% envision panoptica as a valuable tool facilitating in-depth evaluation of
% segmentation methods

% \input{body/2_peakgt/fig_pgt}

% \input{body/2_peakgt/related_work.tex}

% \input{body/2_peakgt/theory.tex}

% \input{body/2_peakgt/quantification.tex}

% \input{body/2_peakgt/examples.tex}

\section{Applications and use cases}
\label{sec:applications}
% The modular architecture of BL allows to create processing pipelines for countless applications.
% The BL tutorial \href{https://github.com/BrainLesion/tutorials}{BL tutorials repository} features examples for multiple use cases, such as metastasis and glioma segmentation.

The modular architecture of \eac{BLS} facilitates the construction of diverse customizable processing pipelines applicable to a wide range of applications.
Exemplified by the \eac{BLS} tutorial's demonstrations of metastasis and glioma segmentation \href{https://github.com/BrainLesion/tutorials}{BL tutorials repository}, this framework lends itself to numerous use cases.

This section subsequently presents a selection of prevalent applications within the neuro- radiology/oncology domain for which \eac{BLS} offers a robust modular platform.

\begin{itemize}
    \item \textbf{Lesion Segmentation} Precise segmentation of lesions in \eac{MRI} of the brain is fundamental for the development of clinically relevant decision-support systems, including those for diagnosis and prognosis.
          The \eac{BLS} provides essential components for segmentation applications, facilitating both the development of new models and the application of pre-trained models for inference.

          Optimizing the performance of segmentation models necessitates meticulous preparation and preprocessing of \eac{MRI} scans.
          The \textit{Medsort} and \eac{BLP} modules within the \eac{BLS} address these important data preparation and preprocessing steps.
          Notably, the standardization of \eac{MRI} sequence tags across datasets, particularly in \eac{mpMRI} analysis, is important.
          This becomes especially salient when DICOM metadata, including sequence tags, have been removed during anonymization procedures following extraction from clinical PACS.
          Following data structuring with \textit{Medsort}, the \eac{BLP} module enables the transformation of data into a format suitable for analysis.
          While the segmentation of certain pathologies, such as meningioma, requires the analysis of whole-brain scans, other lesions, including multiple sclerosis, glioma, stroke, and metastasis, deal with intracranial contents.
          Therefore, skull stripping and defacing procedures can be employed.
          In \eac{mpMRI} analysis, it is frequently observed that not all \eac{MRI} sequences are acquired in a uniform 3D space.
          In fact, variations in anatomical planes (e.g., axial, sagittal, coronal) are common.
          Therefore, resampling and co-registration of all \eac{MRI} sequences for each subject, facilitated by the preprocessing module, are crucial for spatial alignment.
          Furthermore, registration of image scans to standardized atlases, such as SRI24 or MNI152 (or other relevant atlases), transforms individual brain images into a common spatial framework, enabling harmonized analysis of large-scale datasets.
          Additional preprocessing steps, including intensity normalization and bias field correction, can be applied as required by the specific applications.

          Upon completion of preprocessing, the development of new segmentation models can be executed.
          Alternatively, for numerous applications, robust trained models are available for inference.
          Examples include \eac{AURORA} for metastasis, \textit{GlioMODA} and \textit{PeTu} for glioma in adult and pediatric tumor segmentation, and the \textit{BraTS orchestrator}, which provides access to high-performing models from the \eac{BraTS} challenge for various brain tumor entity segmentation and synthesis applications.

          Enhancing the interpretability and reliability of segmentation models can be achieved using the \eac{DQE} module.
          A significant application of \eac{DQE} is the implementation of \eac{QC} and commissioning procedures for segmentation results in large-scale studies. Specifically, \eac{DQE} facilitates the identification of unreliable segmentation masks, eliminating the need for manual inspection of all masks.
          This capability is crucial for the efficient identification and subsequent review and correction of failure cases.

          Last but not least, the \eac{Panoptica} module provides instance-wise quantification metrics, complementing conventional segmentation metrics.
          This is particularly valuable for pathologies such as ischemic stroke or metastasis, which can manifest as numerous non-overlapping instances.
          These metrics enhance the clinical interpretation of model performance.

    \item \textbf{Imaging Biomarkers Studies} Quantitative imaging features, commonly referred to as imaging biomarkers, hold potential for the advancement of personalized medicine.
          However, their clinical applicability depend on robust reproducibility.
          Rigorous validation of reproducibility is important to ensure the reliability and clinical significance of imaging biomarkers, thereby facilitating accurate disease characterization, treatment response monitoring, and treatment outcome prediction.
          Reproducibility, in this context, denotes the consistency of biomarker measurements across variations in conditions, including image acquisition, processing, and analysis algorithms.
          While image acquisition protocols lie outside the scope of this investigation, the \eac{BLS} platform provides essential tools for the processing and analysis of brain \eac{MRI} volumes, irrespective of the feature extraction methodology employed, whether handcrafted, such as radiomics, or \eac{DL} models.
          Within the domain of brain imaging biomarkers, \eac{BLS} can serve as a foundational infrastructure for diagnostic and prognostic applications such as pathology classification \citep{AkbariTumorClassification2024}, radiogenomics analysis \citep{AnahitaRadiogenomics2025}, treatment response prediction \citep{SpeckterMeningiomaResponse2022}, and survival prediction \citep{BakasGBMSurvival2022}.

          Specifically, the \eac{BLP} plays a critical role in enhancing the reproducibility of imaging biomarker studies through the harmonization of \eac{MRI} data.
          For instance, in \eac{mpMRI} analysis, precise co-registration of \eac{MRI} sequences ensures the extraction of features from corresponding anatomical regions across sequences, while registration to an anatomical atlas facilitates feature extraction from homologous anatomical regions across the dataset.
          Beyond preprocessing, accurate segmentation of pathological regions, such as tumors, from preprocessed images defines the \eac{ROI} for handcrafted feature extraction.
          Robust segmentation models, such as those developed within the \eac{BraTS} challenges, can be employed for this purpose.
          Furthermore, these models can be utilized for stability analysis of handcrafted features by segmenting \eac{ROI} from test-retest scans, enabling the evaluation of the reliability and repeatability of defined feature sets.
          Last but not least, the quality-aware loss function implemented within the \eac{DQE} module can be leveraged during the training of end-to-end \eac{DL} models to enhance the interpretability of learned features.

    \item \textbf{Longitudinal Brain \eac{MRI} Analysis} Longitudinal \eac{MRI} studies are essential for understanding dynamic changes in brain structure and function within individual subjects across multiple time points.
          The development of diagnostic, prognostic, predictive, and monitoring models based on such data provides critical insights into brain development, aging trajectories, disease progression, and treatment response.
          However, inherent variability and confounding factors within longitudinal \eac{MRI} datasets can compromise the reliability and generalizability of derived models.
          To mitigate these challenges, \eac{BLS} offers standardized preprocessing and harmonization strategies to minimize inter-scan variations.

          The \eac{BLP} module standardizes input \eac{MRI} data for AI models by spatially aligning individual patient scans to a canonical anatomical atlas.
          Specifically, the SRI24 atlas, employed as the anatomical template, has demonstrated superior consistency across scans acquired from disparate \eac{MRI} scanners, compared to alternative atlases \citep{menze2014multimodal}.
          The subsequent brain extraction step removes non-cerebral tissue, which would otherwise introduce artifacts into downstream analyses.
          Furthermore, intensity normalization, aimed at achieving uniform relative value ranges, is crucial for reliable extraction of handcrafted features \citep{ScalcoIntensityNorm2022}, particularly when quantifying longitudinal changes (e.g., delta radiomics).

          Beyond preprocessing, \eac{BLS} leverages generative AI models within the \textit{BraTS orchestrator} module to harmonize image data.
          It is not uncommon to observe inconsistencies in the availability of \eac{MRI} sequences in longitudinal brain \eac{MRI}, which can severely limit data utility and, in extreme scenarios, necessitate case exclusion.
          Robust generative models, developed within the \eac{BraTS} Synthetic challenge framework, can synthesize realistic and anatomically plausible \eac{MRI} sequences, thereby addressing data incompleteness.

          Last but not least, the accurate tracking of target region changes in longitudinal \eac{MRI} studies necessitates precise segmentation.
          Manual annotation, however, is time-intensive and susceptible to inter- and intra-rater variability.
          Preprocessed and harmonized longitudinal datasets can be employed for the development of longitudinal segmentation models \citep{huangLogitudinalSeg2022} or for inference and prediction using robust, pre-existing segmentation models, such as those integrated within \eac{BLS}.

    \item \textbf{Brain Tumor Growth Modeling} \eac{GBM} exhibits a unique characteristic among cancers, wherein mortality is primarily attributed to localized progression \citep{MetzGBMTherapy2023}.
          The highly infiltrative nature of \eac{GBM} renders conventional \eac{MRI} insufficient for differentiating between surrounding brain tissue and peritumoral edema from non-enhancing tumor infiltration.
          Therefore, the development of models capable of predicting individual tumor spread beyond the \eac{MRI}-visible boundary represents a significant advancement in the diagnosis and treatment of this fatal disease.
          Whether employing biophysical modeling or \eac{DL} methodologies, \eac{mpMRI} data are necessary for deriving parameters through model calibration.

          \eac{BLS} provides necessary modules as infrastructures for brain tumor growth modeling, encompassing preprocessing, segmentation, and synthesis modules.
          The preprocessing module facilitates registration to standardized brain atlases, such as SRI24 or MNI152, enabling robust tissue segmentation into white matter, gray matter, and \eac{CSF}, which are crucial for growth modeling.
          Furthermore, the removal of non-cerebral tissues minimizes artifact contamination during subsequent modeling stages.
          Complementing the preprocessing module, the segmentation module of \eac{BLS} offers accurate \eac{GBM} segmentation algorithms, including \textit{GlioMODA} and models from the \eac{BraTS} package.
          These models accurately segment GBM subregions, such as the active, necrotic, and edematous components, which can serve to estimate the initial coordinate of \eac{GBM} and guide the growth modeling.
          An important requirement for growth modeling is the estimation of the subject's healthy brain state.
          However, pre-diagnostic healthy \eac{MRI} scans are typically unavailable for patients.
          Models developed for the \eac{BraTS} Inpainting task offer potential solutions for reconstructing healthy brain tissue within tumoral regions, facilitating the forward growth of the simulations.

    \item \textbf{Other Applications} The modules described within \eac{BLS} establish an essential infrastructure capable of supporting various advanced \eac{DL}-based applications relevant to brain \eac{MRI} analysis, including but not limited to federated learning \citep{SarthakFL2022}, active learning \citep{ZHANGActiveLearning2021}, and multimodal studies \citep{GuoMultiModal2024}.
          This framework underpins advanced \eac{DL} projects by enabling important operational capabilities.
          These capabilities encompass the standardization and harmonization of large-scale examined data, the identification of potentially unreliable segmentation masks, the enhancement of model interpretability, the achievement of robust segmentation results, and the quantification of model performance via clinically relevant metrics, thereby showcasing the comprehensive support provided by \eac{BLS} for complex brain \eac{MRI} studies.
\end{itemize}

\section{Discussion}
% section is still crude
\label{sec:discussion}
This work presents \eacf{BLS} a flexible and modular framework for constructing robust and reproducible pipelines for brain lesion image analysis.  
The design addresses key challenges in neuroimaging, including heterogeneous \eac{MRI} acquisition protocols, missing modalities, anatomical misalignments, and inconsistent labeling.  
By combining standardized preprocessing, adaptable segmentation workflows, and advanced evaluation methods, the system enables streamlined image analysis suitable for both research and translational applications.

One of the major strengths of the proposed framework is its ability to support the entire neuroimaging workflow—from raw data organization and harmonization to segmentation, synthetic image generation, quality control, and performance evaluation.  
This modular architecture facilitates the construction of tailored pipelines that can adapt to varied scientific or clinical objectives, including diagnostic imaging, biomarker development, and treatment planning.

\textbf{Limitations:} Nevertheless, several limitations must be acknowledged.  
Current segmentation models are optimized for specific lesion types and may not generalize well to other pathologies or imaging domains without retraining.   
Additionally, while the framework is technically robust, its reliance on command-line tools and scripting may limit accessibility for users without programming experience.  
It is also important to emphasize that the framework has not been certified as a medical product and is therefore not intended for unsupervised clinical diagnostic or therapeutic use. 
However, in some settings, clinicians may review and approve algorithmic outputs under their responsibility, allowing \eac{BLS} to support, but not replace, medical decision-making.

\textbf{Outlook:} Looking ahead, several promising directions could extend the scope and usability of the framework.  
Expanding support for additional lesion types, such as ischemic stroke and multiple sclerosis, would enhance its relevance for a broader range of neurological diseases.  
Integrating tumor growth modeling capabilities, through either biophysical simulations or data-driven approaches, could support research on tumor progression and improve individualized treatment planning.  
To improve accessibility and facilitate wider adoption, future work should focus on enabling \eacp{GUI}, for example, via integration with established platforms such as \href{https://www.slicer.org/}{3D Slicer}.  
These enhancements would allow for more intuitive interaction with the pipeline and streamline its integration into clinical and research workflows for broader audiences.

In summary, the presented framework offers a powerful and extensible foundation for brain image analysis.  
Its modularity and emphasis on reproducibility make it well-suited for a wide range of scientific applications.  
Future efforts should aim to expand its clinical scope, lower the technical barrier to entry, and explore pathways for certification and clinical integration.

% \include*{body/9_other/acknowledgment}
% \let\clearpage\relax
\bibliographystyle{plainnat}

\bibliography{references}

\end{document}